\documentclass[lettersize,journal]{IEEEtran}
\usepackage{graphicx}          
\usepackage{amsmath}
\usepackage{tabularx}
\usepackage{amsfonts}
\usepackage{url}
\usepackage[table,xcdraw]{xcolor}
\usepackage{cite}
\usepackage{verbatim}
\usepackage{times,epsfig}
\usepackage{makecell}
\usepackage{psfrag}
\usepackage{subfigure}
\usepackage{stfloats}
\usepackage{setspace}
\usepackage{amssymb}
\usepackage{multirow}
\usepackage{colortbl}
\usepackage[justification=centering]{caption}
\usepackage{fancyhdr}
\usepackage{utfsym}
\usepackage{makecell}
\usepackage{booktabs}
\usepackage{tikz}
\usepackage{hyperref}
\usepackage{algorithmic}
\usepackage[ruled,vlined,linesnumbered]{algorithm2e}

\usepackage{enumitem}
\newlist{steps}{enumerate}{1}
\setlist[steps, 1]{label = Step \arabic*:}

\usepackage{upgreek}
\usepackage{amsthm}

\newenvironment{proof*}[1][\proofname]{{\bfseries #1.}}{\qed}
\newtheorem{thm}{Theorem}

\newtheorem{asmp}{Assumption}

\usepackage{geometry}

\def\BibTeX{{\rm B\kern-.05em{\sc i\kern-.025em b}\kern-.08em
    T\kern-.1667em\lower.7ex\hbox{E}\kern-.125emX}}

\begin{document}
	
	\pagenumbering{arabic}

 \title{On-demand  Quantization for Green Federated Generative Diffusion in Mobile Edge Networks}
\pagestyle{empty}
\thispagestyle{empty}
	\author{\IEEEauthorblockN{Bingkun~Lai\IEEEauthorrefmark{1}, Jiayi~He\IEEEauthorrefmark{1}, Jiawen~Kang\IEEEauthorrefmark{1},Gaolei~Li\IEEEauthorrefmark{2},Minrui~Xu\IEEEauthorrefmark{3},Tao~zhang\IEEEauthorrefmark{4},Shengli~Xie\IEEEauthorrefmark{1},~\IEEEmembership{Fellow,~IEEE}}

    \IEEEauthorblockA{\IEEEauthorrefmark{1}\textit{School of Automation, Guangdong University of Technology, Guangzhou, China}\\
    \IEEEauthorrefmark{2}\textit{School of Electronics Information and Electrical Engineering, Shanghai Jiao Tong University, Shanghai, China}
    \IEEEauthorrefmark{3}\textit{School of Computer Science and Engineering, Nanyang Technological University, Singapore}\\
    \IEEEauthorrefmark{4}\textit{ School of Software Engineering, Beijing Jiaotong University, Beijing, China}
                      }

	\thanks{
    This work was supported by the National Natural Science Foundation of China (NSFC) under Grants No. 62102099, No. U22A2054, the Pearl River Talent Recruitment Program under Grant 2021QN02S643, the Talent Fund of Beijing Jiaotong University under Grant 2023XKRC050, the National Funded Postdoctoral Research Program under Grant GZC20230223, and Guangzhou Basic Research Program under Grant 2023A04J1699, and is also supported by Energy Research Test-Bed and Industry Partnership Funding Initiative, Energy Grid (EG) 2.0 programme, DesCartes and MOE Tier 1 (RG87/22).
    \textit{corresponding author: Jiawen Kang (e-mail: kavinkang@gdut.edu.cn)}}
	}
	\maketitle
	\pagestyle{headings}

	\begin{abstract}
    Generative Artificial Intelligence (GAI) shows remarkable productivity and creativity in Mobile Edge Networks, such as the metaverse and the Industrial Internet of Things. Federated learning is a promising technique for effectively training GAI models in mobile edge networks due to its data distribution. However, there is a notable issue with communication consumption when training large GAI models like generative diffusion models in mobile edge networks. Additionally, the substantial energy consumption associated with training diffusion-based models, along with the limited resources of edge devices and complexities of network environments, pose challenges for improving the training efficiency of GAI models. To address this challenge, we propose an on-demand quantized energy-efficient federated diffusion approach for mobile edge networks. Specifically, we first design a dynamic quantized federated diffusion training scheme considering various demands from the edge devices. Then, we study an energy efficiency problem based on specific quantization requirements. Numerical results show that our proposed method significantly reduces system energy consumption and transmitted model size compared to both baseline federated diffusion and fixed quantized federated diffusion methods while effectively maintaining reasonable quality and diversity of generated data.
	\end{abstract}

	
	\section{Introduction}
         
         As the carrier of content flow, mobile edge networks become essential fundamentals of next-generation applications like Metaverse~\cite{10313945} and the Industrial Internet of Things. 
         The generative models like GAN~\cite{GAN} have demonstrated excellent performance in trajectory prediction~\cite{GeneratingVehicle}, education~\cite{education}, healthcare~\cite{healthcare}, and other scenarios involving the Internet of Things and the Internet of Vehicles. Therefore, more creative generative diffusion models are expected to be deployed in mobile edge networks for next-generation application scenarios such as the 6G communication networks~\cite{10318165} and vehicular metaverses~\cite{VehicularMetaverse},~\cite{luo2023privacy}. Towards deploying the generative diffusion models in mobile edge networks, distributed training schemes based on federated learning called federated diffusions~\cite{FederalDiffusion, EfficientDiffusion}are proposed.  These innovative models enable mobile edge networks to achieve higher productivity and efficiency in next-generation application scenarios.

         During the training phase of federated diffusions, the model needs to be transferred between the server and edge devices at each training step to update the global model~\cite{10044183}.
         This is not a problem when training traditional AI models using federated learning due to their small number of model parameters. However, generative diffusion models are usually large, leading to significant energy expenditure during the federated training process~\cite{huang2023federated}. Therefore, reducing training energy costs is crucial for improving overall operational efficiency in mobile edge networks~\cite{energyAIGC}.
         
          Recent research on federated diffusions has primarily focused on improving their task performance~\cite{FederalDiffusion}. This includes endeavors to elevate the quality and diversity of generated content. However, there is still limited depth and scope in studies that aim to optimize the overall training cost of these models. The authors in~\cite{yang2020energy},~\cite{li2023snowball} studied the problem of energy-efficient resource allocation of FL over wireless communication networks. They derived the energy consumption models for FL based on the convergence rate analysis. The authors in~\cite{meng2022quantized} explored post-training quantization techniques for diffusion models, allowing direct quantization into 8 bits without significant performance degradation, with no substantial decline in performance. However, existing works fail to take into account the substantial training costs of diffusion models or the trade-off between performance and efficiency in the context of complex generative diffusion models~\cite{ho2020denoising}. 
         
         To explore the deployment of a green generative diffusion model in mobile edge networks, we propose a dynamic quantization scheme for transmitting models during federated diffusion training. Firstly, we compress the diffusion models using a quantization scheme before transmission. We then study an energy consumption optimization problem and its solution. The performance of our proposed scheme is evaluated through simulations on the DDPM~\cite{ho2020denoising} model. Our main contributions can be summarized as follows:
         
         \begin{itemize}
	    \item We design a new and environmentally friendly federated generative diffusion framework that utilizes a dynamic method for parameter quantization and training in mobile edge networks. 
		\item We formulate an optimization problem for resource allocation in dynamic quantized federated diffusion, aiming to minimize total energy consumption while maintaining commendable performance.
		\item Numerical results demonstrate the effectiveness of our proposed method compared to other baseline methods, particularly in terms of energy efficiency and sample quality.
	\end{itemize}

        The structure of the paper is organized as follows. The system model and the proposed on-demand quantized federated diffusion framework are introduced in section ~\ref{section 2}. Next, we study the energy efficiency optimization problem in section ~\ref{sec 3}. Finally, We show the simulation results in section ~\ref{sec 4} and discuss the conclusion and future work in section ~\ref{sec 5}.

 \section{System Model}\label{section 2}

\begin{figure}[t]
    \centering  
    \includegraphics[width=0.55\textwidth]{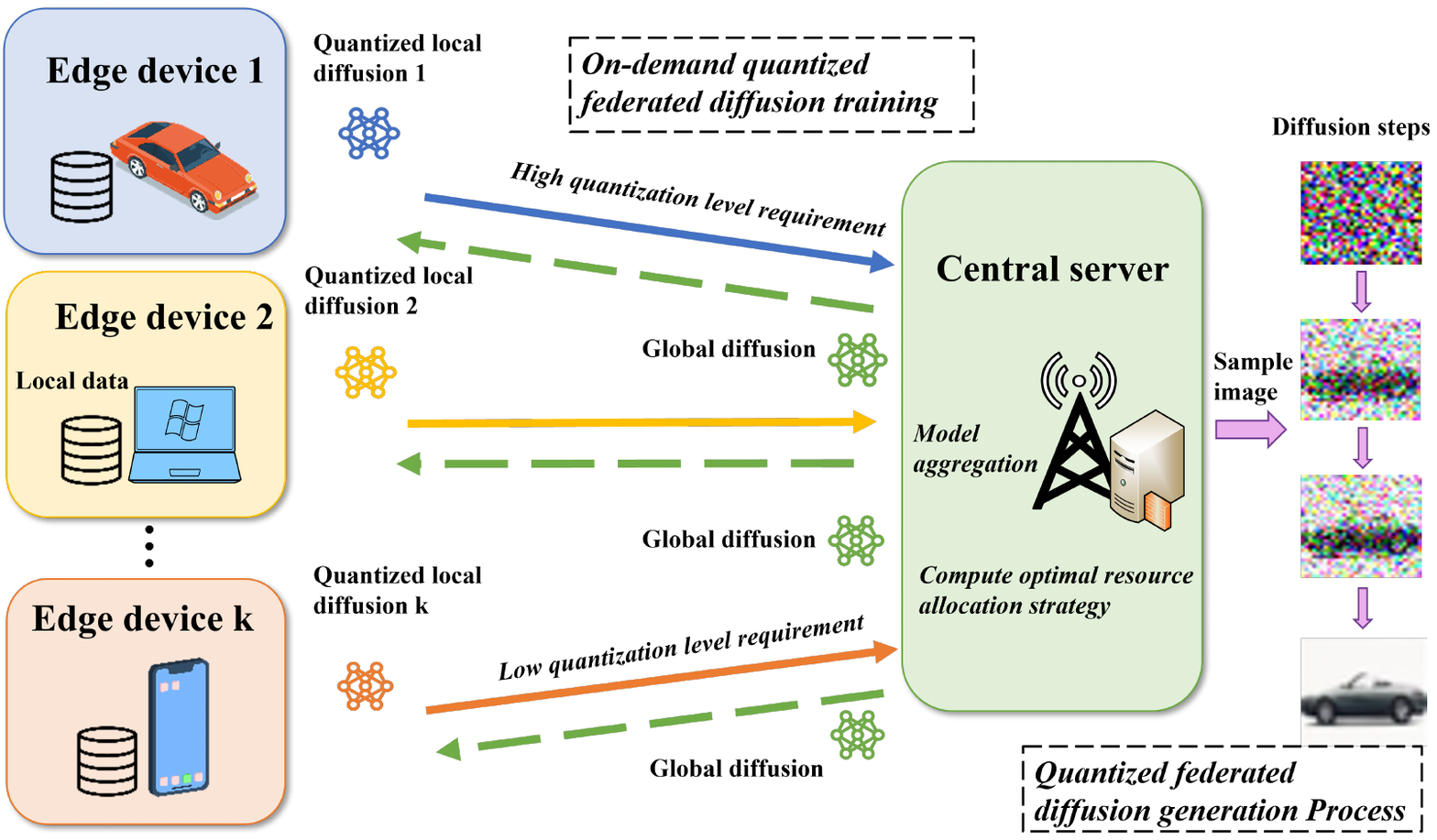}   
    \caption{On-demand quantized federated diffusion framework}
    \label{fig1}
\end{figure}

    As shown in Fig.~\ref{fig1}, we consider a mobile edge network scenario where a central server and $k$ edge devices collaborate to train a diffusion model using federated learning. Given the inherent characteristics of large diffusion models, training them in federated learning scenarios can be exceptionally energy-intensive. A promising model compression method named stochastic quantization~\cite{chen2022energy} is implemented prior to the transmission of model parameters from edge devices to the edge server for aggregation, this is done to mitigate the transmission costs of the training process. Additionally, we take into account the variable quantization level needs of edge devices, ensuring the flexibility of quantization to accommodate different device requirements. Furthermore, considering the heterogeneous nature of edge devices and their varying resource capacities, an energy optimization problem is formulated to further minimize the energy consumption during federated diffusion training. After the efficient training is done, the server could utilize the final global diffusion model for efficient and high-quality content generation. The learning process for each round of iteration is as follows:
    
    \begin{itemize}
          \item \textit{Step 1}: Given different quantization requirements, the central server determines the optimal strategy for each edge device to balance computing and communication resources based on the resource status of different devices. 
          \item \textit{Step 2}: The edge devices then perform a local diffusion computation and transmission according to the optimal strategy.
          \item \textit{Step 3}: After receiving all local diffusions from edge devices, the central server uses an aggregation scheme (such as Fedavg~\cite{mcmahan2017communication}) to unite the local diffusion into a new
          global diffusion and send it back to the edge devices for next round of training.
        \end{itemize}
    
    Hereinafter, we introduce the concept of quantization, which is a promising method for compressing neural networks. Stochastic quantization can be efficiently used in the federated learning process to significantly reduce energy consumption while maintaining minimal impact on model performance. To minimize the cost of transmitting a comparatively large model, as in Fig.~\ref{fig1}, we propose quantizing the local diffusion model before uploading it to the server, since the resources of edge devices are often limited. To train the federated diffusion model with a quantization scheme, we first define the stochastic quantization function as $Q(\cdot)$. Given the local diffusion weight ${\boldsymbol w}_k$, the quantized weight can be expressed as $\hat{\boldsymbol w}_k = Q({\boldsymbol w}_k)$. Let $|{\boldsymbol w}^{[n]}_k|$ denotes the absolute value of element in ${\boldsymbol w}_k$, The stochastic quantization function is defined as 

\begin{equation}
Q({\boldsymbol w}_k)= a\cdot sign({\boldsymbol w}_k)\cdot
\begin{cases}
    q^{l+1} & \text{w.p. $\frac{|{\boldsymbol w}^{[n]}_k|-aq^{l}}{a(q^{l+1}-q^l)}$}\\
    q^l &  \text{w.p. $\frac{aq^{l+1}-{|{\boldsymbol w}^{[n]}_k|}}{a(q^{l+1}-q^l)}$} \\
\end{cases}
\end{equation}
Here, $a$ is the scale factor and $sgn(\cdot)$ denotes the sign function which represents the sign of ${\boldsymbol w}_k$. Moreover, $[q^l,q^{l+1}]$ is the quantization interval such that for any ${|\boldsymbol w}^{[n]}_k|$ there exists $\frac{|{\boldsymbol w}^{[n]}_k|}{a} \in [q^l,q^{l+1}]$. With the given quantization level $L_k$, $q^l$ can be calculated as
\begin{equation}
\begin{split}
q^l = \frac{l(\boldsymbol w^{max}_k-\boldsymbol w^{min}_k)}{a(q^{l+1}-q^l)}+ \frac{\boldsymbol w^{min}_k}{a}
\end{split}
\end{equation}
Where $\boldsymbol w^{max}_k$ and $\boldsymbol w^{min}_k$ represent the maximum and minimum value of non-zero element $|{\boldsymbol w}^{[n]}_k|$ , respectively.

Next, we formulate the computation and communication models of the proposed scheme. Let $f_k$ represent the computation frequency of local client $k$, and $D_k$ denotes the data size of the local dataset. The computation time for training diffusion model is expressed by 
\begin{align}
\begin{split}
{T^{cmp}_k}= \frac{I_k{D_k}{C}}{f_k}
\end{split}
\end{align}
where $I_k$ and $C$ denote the local iteration times in each communication round and the workload of local diffusion training, respectively. Following that, given the energy coefficient $\tau_k$, the energy consumption of client $k$ is estimated by
\begin{align}
\begin{split}
{E^{cmp}_k} = {\tau_k}{{f_k}^2{I_k}{D_k}{C}}
\end{split}
\end{align}
In the distributed diffusion setting, each edge device uploads the local diffusion model in order to generate a better global model. Moreover, the local diffusion model is quantized as $\hat{\boldsymbol w}_k$ for efficiency improvement. To this end, we adopt a frequency division multiple access (FDMA) transmission scheme~\cite{myung2006single} for quantized local diffusion transmission. Therefore, with the transmission power $P_k$, the uplink transmission rate of client $k$ is deduced by
\begin{align}
\begin{split}
{r_k}={B\log_2{(1+\frac{|h|^2 d^{-\eta} P_k}{B N_0})}}
\end{split}
\end{align}
Here, $B$ and $N_0$ denote the bandwidth and noise power-spectral-density each while $d$ corresponds to the distance between the client and server. Meanwhile, $h$ and $\eta$ represent the Rayleigh
channel coefficient and pathloss exponent, separately. Subsequently, given the updated model size $M_k$, and quantization level $L_k$, the time spent by client $k$ to transmit the local model to the server is
\begin{align}
\begin{split}
{T^{com}_k}=\frac{M_k L_k}{r_k} 
\end{split}
\end{align}
Thus, the corresponding energy consumption is calculated by 
\begin{align}
\begin{split}
{E^{com}_k}= P_k{T^{com}_k}
\end{split}
\end{align}

\section{Energy Efficiency Optimization}\label{sec 3}
     
        
        Before the discussion of the energy efficiency problem of the proposed methods started, similar to the work in~\cite{li2023snowball}, we first present the following assumption and theorem. Where $\delta_k$ represents the unique bound demand of various edge devices, a smaller $\delta_k$ indicates that the resource of edge device $k$ is relatively insufficient, which leads to a lower quantization level strategy:

        \begin{asmp}
        \label{asmp:1}
        the expectation of the square norm of the local weight uploaded by edge devices is bounded: for any uploaded weight, ${\mathbb E}\|{\boldsymbol w}_k\|^2 \le \delta_k$.
        \end{asmp}
        
        \begin{thm}
        Based on Assumptions \ref{asmp:1}, the square of local weight quantization error ${\Delta}_k$ is bounded by:
        \begin{align}
        \begin{split} {\Delta}_k={\mathbb E}\| {\boldsymbol w}_k - \hat{\boldsymbol w}_k\|^2 \le \frac{\delta_k}{2L^2_k}.
        \end{split}
        \label{eq:thm2}
        \end{align}
        \label{thm:2}
        \end{thm}
Through this theorem, we can easily obtain the corresponding quantization levels for each heterogeneous edge device at different demands, thereby further constructing our energy consumption optimization model, which we will discuss in detail in the next section.

\subsection{Problem Formulation}

As a consequence of variations in resource capabilities among edge devices, variability exists in the required quantization level demands. In simpler terms, each device has its distinct upper bound for quantization error. Leveraging Theorem 1, we establish the energy minimization problem within the confines of this quantization error constraint as follows:

\begin{subequations}\label{eqn:pbm-1}
\begin{align}
({\text{P1}}) && \min \limits_{P_k,f_k,L_k}( E^{cmp}_k & + E^{com}_k)    \quad   &\tag{\ref{eqn:pbm-1}}\\
{\text{subject to:}} &&  T_k^{cmp} + T_k^{com}  \le &\;{T_k^{max }}, \forall k \label{eqn:p1-ctr-1}\\
{}&& {\mathbb E}\| {\boldsymbol w}_k - \hat{\boldsymbol w}_k\|^2 \le  &\;\frac{\delta_k}{2L^2_k}, \forall k \label{eqn:p1-ctr-2}\\
{}&& P^{min}_k \le P_{k} \le  &\; P^{max}_k, \forall k \label{eqn:p1-ctr-3}\\
{}&&  f_k^{min} \le f_{k} \le &\; f_k^{max}, \forall k \label{eqn:p1-ctr-4}
\end{align}
\end{subequations}

With the unique ${\delta}_k$ given by the different bound requirements of local edge devices, we can always obtain the optimal solution $L^*_k = \sqrt{\frac{{\delta}_k}{2{\Delta}_k}}$ for the optimization problem. As a result, we simplify P1 as
\begin{subequations}\label{eqn:pbm-2}
\begin{align}
({\text{P2}}) && \min  \limits_{P_k,f_k}( E^{cmp}_k & + E^{com}_k)   \quad   &\tag{\ref{eqn:pbm-2}}\\
{\text{subject to:}} &&  T_k^{cmp} + T_k^{com}  \le &\;{T_k^{max }}, \forall k &\label{eqn:p1-ctr-2}\\
{}&& P^{min}_k \le P_{k} \le  &\; P^{max}_k, \forall k \label{eqn:p1-ctr-3}\\
{}&&  f_k^{min} \le f_{k} \le &\; f_k^{max}, \forall k \label{eqn:p1-ctr-4}
\end{align}
\end{subequations}
Following that, we transform P2 into a more tractable form by introducing two intermediate variables $\theta_k > 0$ and $\pi_k > 0$. Moreover, we let $\theta_k$ and $\pi_k$ represent the weight factors of maximum time budget for client $k$ such that   
\begin{align}
\begin{split}
{\theta_k}{T_k^{cmp}} =  \frac{I_k{D_k}{C}}{f_k} \text{,}\\ {\pi_k}{T_k^{com}} = \frac{M_k \log_2{(L_k)}}{r_k} 
\end{split}
\end{align}
Here, the lower bound of $\theta_k $ and $\pi_k $ can be easily acquired given the optimal $L_k$ 
\begin{align}
\begin{split}
{\theta_k^{min}}=\frac{I_k{D_k}{C}}{f_k^{max} T_k^{max}} \text{,}\\
{\pi^{min}_k} = \frac{M_k \log_2{(L^*_k)}}{{B T_k^{max}\log_2{(1+\frac{|h|^2 d^{-\eta} P^{max}_k}{B N_0})}}} 
\end{split}
\end{align}
Furthermore, the total energy consumption of client $k$ during the fine-tuning process can be rewritten in the following form
\begin{align}                       
\begin{split}
{E_k} &= E^{cmp}_k + E^{com}_k \\
&= \frac{\tau  I^3_k D^3_k C^3_k}{\theta_k^2 (T_k^{max})^2} + \frac{N_0 B T^{max}}{\lvert h \rvert^2 d^{-\eta}} \big(2^{\frac{M_k \log_2{(L^*_k)}}{\pi_k B T_k^{max}}}-1\big)  
\end{split}
\end{align}
Thus, we convert problem P3 into the following form
\begin{subequations}\label{eqn:pbm-3}
\begin{align}
(\text{P3}) &&   \min  \limits_{\theta_k,\pi_k} E_k   \quad \thinspace  &\tag{\ref{eqn:pbm-3}}\\
{\text{subject to:}} &&  \theta_k+\pi_k =1, \forall k  &&    &\label{eqn:p1-ctr-2}\\
{}&&  \;\theta_k^{min} \le  \theta_k, \forall k \label{eqn:p3-ctr-2}\\
{}&&  \;\pi_k^{min}  \le  \pi_k, \forall k \label{eqn:p3-ctr-3}
\end{align}
\end{subequations}
Through this basic form,  we can readily acquire the numerical solution for the original problem. In the next subsection, we will present the solution to address the current matter.
\subsection{Solution}
It can be easily proved that problem P3 is a convex problem, which can be effectively solved by applying the Karush-Kuhn-Tucker (KKT) conditions~\cite{boyd2004convex}. With the optimal energy optimization solution, we can decide the final resource allocation scheme for the federated diffusion. The Lagrange function of P3 is as follows:
\begin{align}
\begin{split}
{\boldsymbol{L}}(P_k,f_k,{\nu}_k,\zeta_k^{\theta},\zeta_k^{\pi}) = \frac{\tau  I^3_k  D^3_k C^3_k}{\theta_k^2 (T_k^{max})^2}\\
+ \frac{N_0 B T_k^{max}}{\lvert h \rvert^2 d^{-\eta}} \big(2^{\frac{M_k \log_2{(L^*_k)}}{\pi_k B T_k^{max}}}-1\big) +{\nu}_k(\theta_k + \pi_k-1)\\
+ \zeta_k^{\theta}(\theta_k^{min}-\theta_k)+\zeta_k^{\pi}(\pi_k^{min}-\pi_k)
\end{split} 
  \end{align}
Here, ${\nu}_k$ is the equality Lagrange multiplier associated with equality constraint (14a), while ${\zeta}_k^{\theta}$ and $\zeta_k^{\pi}$ denote the inequality Lagrange multiplier for constraints (14b) and (14c), respectively. In order to accomplish optimality for problem P3, we derive the necessary equations from the Lagrange function as follows:
\begin{equation}
 \left\{
\begin{array}{c}
{\text{constraint (14a)-(14c)}}\\

\frac{2\tau I^3_k  D^3_k C^3_k}{\theta_k^3 (T_k^{max})^2} + {\nu}_k -\zeta_k^{\theta}= 0\\
\frac{N_0 B T_k^{max}}{\lvert h \rvert^2 d^{-\eta}}\left(2^{\frac{M_k \log_2{(L^*_k)}}{\pi_k BT_k^{max}}}-1 - \frac{\ln{(2)}M_k \log_2{(L^*_k)}}{\pi_k B T_k^{max}}\right)+{\nu}_k -\zeta_k^{\pi} = 0\\
\zeta_k^{\theta}(\theta_k^{min}-\theta_k)-\zeta_k^{\pi}(\pi_k^{min}-\pi_k) = 0\\

\end{array} \right.
\end{equation}
Based on Eqns 16,  there exist two cases that satisfy constrain (14b) concerning variable $\theta_k$. If $\theta_k > \theta_k^{min}$, the optimal solution of $\theta^*_k$ is obtained by \\
\begin{align}
\begin{split}
\theta^*_k = \sqrt[3]{\frac{2\tau I^3_k  D^3_k C^3_k}{\nu_k(T_k^{max})^2}}
\end{split}
\end{align}
Otherwise, we always have $\theta^*_k = \theta_k^{min}$. Similarly to $\theta_k$, when $\pi_k>\pi_k^{min} $, the optimal $\pi^*_k$ is acquired the same way. Given the equality Lagrange multiplier $\nu_k$, we have\\
\begin{align}
\begin{split}
\Phi(\pi^{0}_k)=\frac{N_0 B T_k^{max}}{\lvert h \rvert^2 d^{-\eta}}(2^{\frac{M_k \log_2{(L^*_k)}}{\pi^{0}_k B T_k^{max}}}-1- \frac{\ln{(2)}M_k \log_2{(L^*_k)}}{\pi^{0}_k B T_k^{max}})\\
+{\nu}_k = 0 
\end{split}
\end{align}

Where $\pi^{0}_k$ is the zero point of function $\Phi(\pi_k)$. In general, the optimal solution of $\theta_k$ and $\pi_k$ can be acquired by 
\begin{align}
\begin{split}
\theta_k = {\max}\{\sqrt[3]{\frac{2\tau I^3_k  D^3_k C^3_k}{\nu_k(T_k^{max})^2}},\theta^{min}_k\},
\pi_k = {\max}\{\pi^{0}_k,\pi^{min}_k\}
\end{split}
\end{align}
It's worth mentioning that seeking the problem's optimal solution directly can be rather intricate, which is why we employed binary search to find the optimal strategy for the Lagrange multiplier $\nu_k$.  Utilizing the most favorable Lagrange multiplier value, the optimal approach for variables $\theta_k $ and $\pi_k$ are computed based on (19). To be specific, given the searching range of $\nu_k$ and error tolerance $\lambda$, the optimal $\nu_k$ is obtained with the constraint (14a). Additional and more detailed information is provided in Algorithm 1. Finally, the overall algorithm of the proposed method is shown as algorithm 2.

\begin{algorithm}
\caption{Binary Search}
\label{alg2}
\SetAlgoLined
\SetKwInput{KwInput}{Input}
\SetKwInput{KwOutput}{Output}
\SetKwRepeat{KwRepeat}{repeat}{until}
\KwInput{$\nu^{min}_k$,$\nu^{max}_k$,$\pi^{min}_k$,$\pi^{max}_k$,and $\lambda$.}
\KwOutput{The optimal Lagrange multiplier $\nu^*_k$.}
\While{$|\nu^{max}_k - \nu^{min}_k| \le \lambda$}{
    $\nu_k =(\nu^{max}_k + \nu^{min}_k)/2$\;
    Calculate $\theta^*_k $\;
    Search for $\pi^*_k$\;
    if $\theta^*_k +\pi^*_k \le 1$ then $\nu^{max}_k = \nu^*_k$ else $\nu^{min}_k = \nu^*_k$ \;
    }
\Return $ \nu^*_k$\  \\
\While{$|\pi^{max}_k - \pi^{min}_k| \le \lambda$}{
    $\pi_k =(\pi^{max}_k + \pi^{min}_k)/2$\;
    Calculate $\Phi(\pi_k) $\;
    if $\Phi > 0$ then $\pi^{max}_k = \pi^*_k$ \
    else $\pi^{min}_k = \pi^*_k$ \;
    }
\Return $ \pi^*_k$\  \\
\label{alg:train}
\vspace{-1pt}
\end{algorithm}

\begin{algorithm}
\caption{Quantized Federated Diffusion}
\label{alg2}
\SetAlgoLined
\SetKwInput{KwInput}{Input}
\SetKwInput{KwOutput}{Output}
\SetKwRepeat{KwRepeat}{repeat}{until}
\KwInput{pre-trained model ${\boldsymbol w}^0$;  variance schedule $\{\beta\}$; iteration $I$;  sample step $T$; error bound $\delta_k$.}
\KwOutput{global model ${\boldsymbol w}^I$.}
\For{$i=0$ to $I$}{
    $K ~\leftarrow$ Select $K$ devices from edge devices pool\;
    Calculate the optimal resource allocation strategy \
    based on Algorithm 1\;
    \For{$k$ in $K$ parallel}{
        Initialize local model ${\boldsymbol w}^i_k$ by ${\boldsymbol w}^i$\;
        A mini-batch original images $x_0$ in local dataset $D_k$\;
        $t \sim Uniform(\{ 1,..., T \})$\;
        $\epsilon \sim \mathcal N (0, I)$\;
        Diffuse $x_0$ to $x_t \approx \epsilon$ by:
        $x_t = \sqrt{ \overline{\alpha}_t}x_0+\sqrt{1-\overline{\alpha}_t}\epsilon$\;
        Take the gradient decent step by minimizing:
        $||\epsilon -F_{{\boldsymbol w}^i_k}(x_t,t)||^2$\;
        Quantized ${\boldsymbol w}^i_k$ based on stochastic quantization\;
        Upload the quantized model $\hat{\boldsymbol w}^{i+1}_k$ to the server.
    }
}
\label{alg:train}
\vspace{-1pt}
\end{algorithm}

\section{Numerical Results}\label{sec 4}
\subsection{Simulation Settings}
    To simulate the practical case of federated diffusion in mobile edge networks, we fine-tuned the pre-trained DDPM~\cite{ho2020denoising} on CIFAR10~\cite{cifar} using the GTSRB~\cite{gtsrb} dataset. The dataset is divided into 10 subsets for 10 edge devices to perform federated learning. We fine-tuned the federated model with 1000 epochs for performance evaluations. The sample steps are set as 1000 for image generation.  For computation and communication hyper-parameters \{${I_k},{D_k},{C},{f_k^{max}},{\tau_k}$\} and \{$B,{|h|^2},d,{\eta},{N_0},{M_k}$\}, the default settings are 
    \{$1, 512, 3.25$MCycles$, 10^9, 10^{-26}$\} and \{$50$MHz$, 0.001$W$, 45$m$, 3.76, -95$dbm/MHz$, 37$M\}.
    
\subsection{Performance Evaluations}
\begin{figure}[h]
    \centering  
    \includegraphics[width=0.51\textwidth]{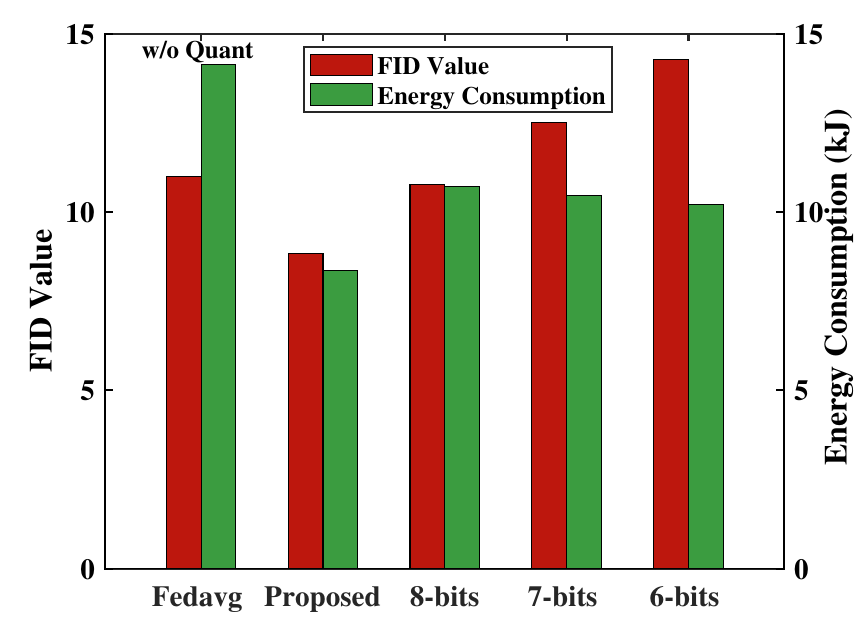}   
    \caption{FID performance and energy consumption of different schemes}
    \label{fig2}
\end{figure}

Fig.~\ref{fig2} illustrates the performance and energy consumption of the proposed algorithm. We employed the Fréchet Inception Distance (FID)~\cite{rombach2022high} as an evaluative metric for assessing the quality of the images generated by the model. A lower FID value indicates a higher degree of similarity between the distribution of the generated dataset and that of the original dataset. To enhance the precision of our evaluation concerning the quality of the generated dataset, we ensured that the number of generated datasets equaled the number of original images. Remarkably, our proposed methodology takes into consideration distinct quantization error constraints customized for heterogeneous edge devices. Subsequently, an energy minimization problem is optimized. The range of quantization levels spans from 6 bits to 8 bits, a range normally associated with a substantial reduction in energy consumption while concurrently upholding commendable performance. We conduct a comparative analysis between the baseline method, Fedavg, in addition to the fixed quantization methods employing 6-bits, 7-bits, and 8-bits quantization levels. It is evident from our results that our approach surpasses the more economical 8-bits quantization scheme in terms of both performance and cost-efficiency. It is essential to note that the compared methods did not specifically optimize for energy consumption, utilizing only $50\%$ time budget for computation and another $50\%$ for communication by default.

\begin{figure}[h]
    \centering  
    \includegraphics[width=0.5\textwidth]{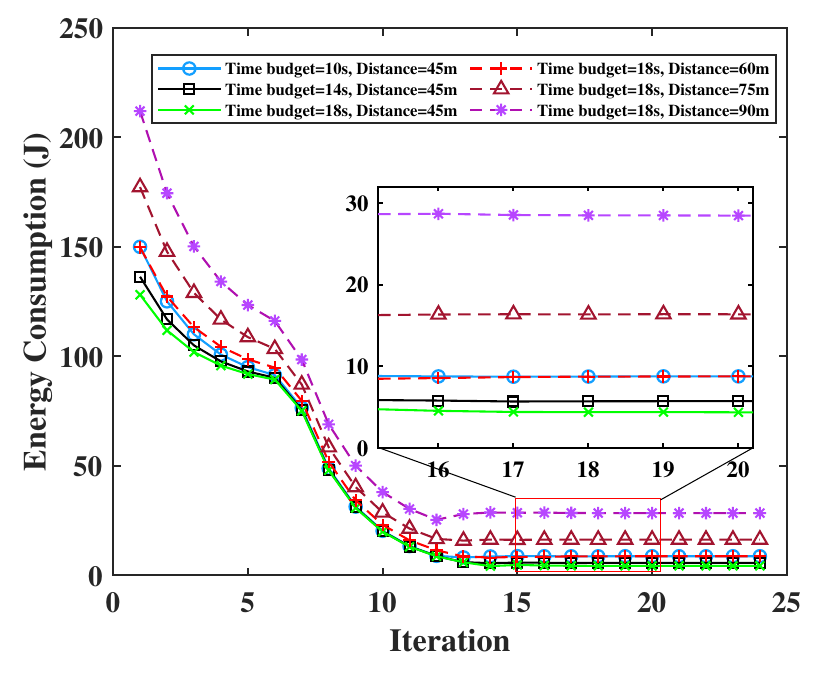}   
    \caption{Convergence of proposed binary search algorithm}
    \label{fig3}
\end{figure}

\begin{figure}[t]
    \centering  
    \includegraphics[width=0.5\textwidth]{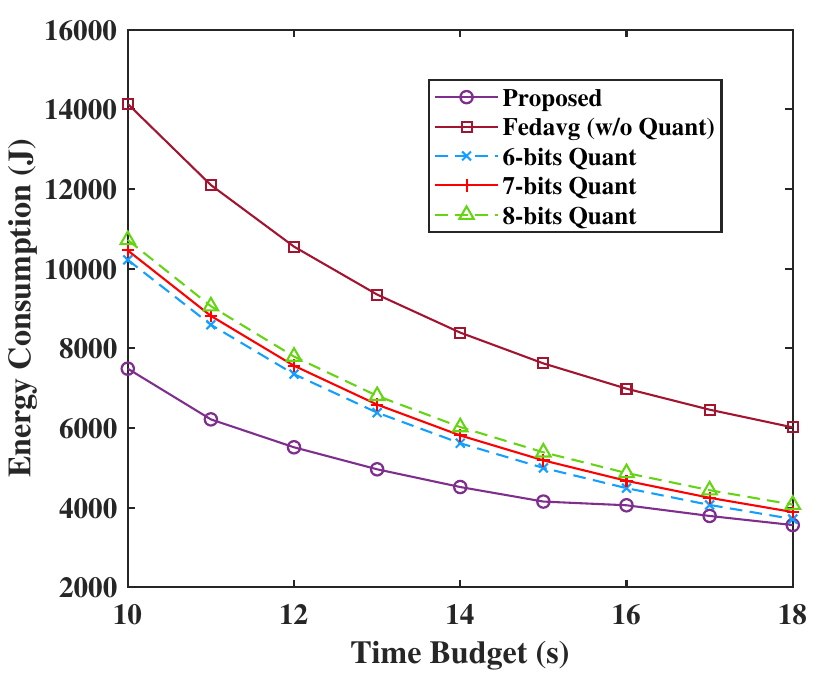}   
    \caption{Energy cost vs. Time budget}
    \label{fig4}
\end{figure}

Fig.~\ref{fig3} illustrates the successful convergence achieved by our proposed binary search algorithm in addressing the energy optimization problem we have established. Specifically, provided various quantization level requirements, the optimal solution for minimizing the energy cost is determined after about 20 searching iterations. Furthermore, a study was conducted to examine the impacts of two different hyperparameters, the time budget and the distance of communication. It can be clearly seen that the proposed method can converge well under different settings. As the allotted time budget reduces, the requisite energy consumption by the system escalates. In parallel, with an augmentation in the distance of communication between edge devices and central servers, there is a concurrent amplification in energy expenditure.
Fig.~\ref{fig4} presents a comparative analysis of our method alongside other approaches across various time budget ranges. As observed, with an increase in the allotted time budget ranging from 13s to 18s, the system's energy costs diminish, and our solution consistently outperforms the baseline approach. This indicates that our method can adapt to parameter settings within certain ranges.

\section{Conclusion and Future Work}\label{sec 5}
    In this paper, we first design a dynamic quantized federated diffusion training considering each edge device's demand. Subsequently, our study turns towards addressing the challenge of energy efficiency, taking into account the unique constraint imposed by quantization demand. Our simulation results demonstrate that our proposed method outperforms both the baseline federated diffusion approach and fixed quantized federated diffusion in substantially reducing system energy consumption and transmitted model size. Remarkably, this reduction is achieved without compromising the reasonable quality and diversity of the generated data, underscoring the effectiveness of our approach.
    
    To achieve the benefits of efficient federated generative diffusion, there still exist several open and challenging issues. For distributed diffusion models, the matter of proficient sampling remains an unsolved problem, primarily due to the distinctive characteristics inherent to diffusion itself. Diverging from conventional AI models, the interference phase of diffusion entails a substantial energy outlay, particularly in the denoising sampling steps. This heightened energy consumption may be deemed unmanageable for certain edge devices. Consequently, further study is imperative to enhance the efficiency of sampling within the context of diffusion, particularly in the domain of distributed edge intelligence scenarios.
\bibliographystyle{ieeetr}
\bibliography{ref}

\end{document}